# Enhancing NLP Robustness and Generalization through LLM-Generated Contrast Sets: A Scalable Framework for Systematic Evaluation and Adversarial Training

Hender Lin

Department of Computer Science, College of Natural Sciences,
The University of Texas at Austin
hlin01@utexas.edu

## Abstract

Standard NLP benchmarks often fail to capture vulnerabilities stemming from dataset artifacts and spurious correlations. Contrast sets address this gap by challenging models near decision boundaries but are traditionally labor-intensive to create and limited in diversity. This study leverages large language models to automate the generation of diverse contrast sets. Using the SNLI dataset, we created a 3,000-example contrast set to evaluate and improve model robustness. Fine-tuning on these contrast sets enhanced performance on systematically perturbed examples, maintained standard test accuracy, and modestly improved generalization to novel perturbations. This automated approach offers a scalable solution for evaluating and improving NLP models, addressing systematic generalization challenges, and advancing robustness in real-world applications.

## 1   Introduction

Advancements in natural language processing (NLP) have been driven by standardized benchmarks like the Stanford Natural Language Inference (SNLI) dataset. While these benchmarks have enabled significant progress, they often fail to assess model robustness in real-world scenarios. High performance on standard test sets can obscure vulnerabilities, as models frequently exploit spurious correlations and dataset artifacts rather than demonstrating true linguistic understanding. This underscores the need for robust evaluation methods to ensure the reliability and applicability of NLP models across diverse contexts.

Contrast sets, introduced by Gardner et al. (2020), address this need by systematically perturbing existing datasets. By focusing on examples near the decision boundary, contrast sets challenge models with instances that demand deeper semantic understanding and careful reasoning. However, their manual curation is labor-intensive, limiting scalability and linguistic diversity. Automated methods, such as those by Lie et al. (2020), improve scalability but often lack the linguistic richness required for comprehensive robustness testing. Large language models (LLMs) offer a promising solution, combining scalability with the ability to generate linguistically diverse and systematic contrast sets.

In this study, we explore the potential of LLMs—specifically Google Gemini 1.5 Pro—to automate the creation of large-scale contrast sets for robust evaluation and adversarial training. Using the SNLI dataset as a case study, we generate a 3,000-example contrast set with balanced label perturbations across six categories, including entailment-to-neutral and neutral-to-contradiction shifts. These contrast sets serve as tools to evaluate model performance and investigate whether adversarial training with this data enhances robustness and generalization to unseen perturbations.

Specifically, we address the following research questions:

1. Can large-scale, LLM-generated contrast sets effectively evaluate NLP model robustness and expose their limitations?

2. Does adversarial training with these contrast sets improve model performance on both original and systematically perturbed datasets?

3. How does adversarial training with these contrast sets impact generalization to unseen perturbations at the decision boundaries?

By fine-tuning the ELECTRA-small model on a combined dataset of the original SNLI training data and the LLM-generated contrast set, we observe significant improvements in robustness without compromising performance on standard benchmarks. Furthermore, the model demonstrates enhanced generalization to unseen perturbations, providing valuable insights into the utility of adversarial training with LLM-generated data. Leveraging the scalability and linguistic diversity offered by LLMs, this study presents a novel framework for robust evaluation and systematic generalization testing of NLP systems.

## 2 Methodology

### 2.1 Contrast Set Generation

We started by constructing a large-scale contrast set derived from the SNLI dataset. This process involved systematically perturbing 500 examples from each label class (entailment, neutral, contradiction), yielding six distinct types of label shifts and a total of 3,000 contrast set examples. The steps for generating this contrast set are outlined in the following subsections.

**Data Selection** To ensure balanced coverage across all label shifts, we randomly sampled 500 examples from each label class within the SNLI test set.

**Automated Generation** Using the Gemini 1.5 Pro API through Google AI Studio, we systematically applied perturbations to the hypotheses. These perturbations were designed to generate linguistically valid transformations that maintained semantic relevance while introducing the intended label shifts. Carefully crafted prompts guided the model in achieving these transformations.

**Prompt Design** Each prompt was meticulously designed to ensure semantic correctness and to minimize deviation from the original hypothesis. For example, the prompt for converting an entailment hypothesis into a contradiction was:

*"Modify the hypothesis so that it directly contradicts the premise. Make the minimal necessary changes to create an explicit contradiction, ensuring the topic and language deviate as little as possible from the original. The contradiction must be obvious and leave no room for ambiguity.*

*Premise: {premise}*

*Original hypothesis (entails): {hypothesis}*

*Provide only the revised hypothesis that contradicts the premise."*

**Validation Process** To verify the quality of the generated examples, 10% of the contrast set was randomly selected for manual review. This review confirmed semantic accuracy and alignment with the intended label shifts, demonstrating the reliability of the outputs. While we initially considered validating label shifts using a pre-trained natural language inference (NLI) model, the manual review process showed that Gemini's generation consistently produced high-quality results. As a result, automated validation was deemed unnecessary, preserving efficiency without compromising rigor.

### 2.2 Augmented Training with Contrast Set

To evaluate the impact of adversarial training with LLM-generated contrast sets, the original SNLI training set (550,000 examples) was augmented with the generated contrast set (3,000 examples). This combined dataset allowed the model to learn from both standard examples and systematically perturbed ones, enhancing its ability to handle linguistic variations at decision boundaries. Details of the training procedure, including hyperparameters and implementation code, are available in the project's GitHub repository.[1]

### 2.3 Evaluation Framework

The evaluation framework assessed the model's robustness and generalization by comparing its performance across three datasets: the original test set, the original contrast set, and a new contrast set containing unseen perturbations. The original test set provided a baseline for measuring overall performance on standard examples. The original contrast set evaluated the model's capability to handle systematic perturbations both before and after adversarial training, while the new contrast set assessed its ability to generalize to novel challenges. Accuracy was the primary metric, offering a direct measure of performance across datasets, both overall and segmented by label type.

---

[1] See code at https://github.com/hlin01/NLP-Final-Project

Additionally, error analysis categorized performance by label shifts, leveraging the substantial size of the contrast set (33% of the original test set) to gain detailed insights into the model's behavior when confronted with systematically perturbed examples. This structured and comprehensive framework ensured a thorough evaluation of the model's improved robustness and its capacity for generalization.

# 3 Results

## 3.1 Baseline Performance

To establish a baseline, we fine-tuned the ELECTRA-small model on the original SNLI training set and evaluated its performance across three datasets: the standard SNLI test set, an LLM-generated contrast set (denoted as the "original contrast set"), and a second LLM-generated contrast set (denoted as the "new contrast set"). On the standard SNLI test set, the model achieved an accuracy of 89.0%, demonstrating strong generalization capabilities on unperturbed data. However, on the original contrast set, performance dropped to 83.2%, indicating a reliance on dataset artifacts and difficulty in handling systematic perturbations. On the new contrast set, which featured novel perturbations, accuracy declined to 85.0%. These results highlight the value of contrast sets in exposing latent model weaknesses and dependencies.

## 3.2 Adversarial Training Improvements

Fine-tuning the model on a combined dataset—consisting of the original SNLI training set and the original contrast set—led to marked improvements in robustness compared to the baseline performance. On the standard SNLI test set, the model achieved an accuracy of 89.1%, a marginal increase of +0.1%, indicating that adversarial training did not compromise performance on in-distribution examples. On the original contrast set, accuracy rose significantly to 90.6% (+7.4%), and on the new contrast set, accuracy increased to 87.9% (+2.9%). These gains are noteworthy, particularly given that adversarial examples comprised only 0.5% of the total training data. Balancing adversarial and original data was critical to achieving these improvements, allowing the model to benefit from challenging examples without overfitting. However, it is likely that increasing the proportion of adversarial data would yield even greater performance gains.

## 3.3 Generalization to Novel Perturbations

The new contrast set, designed to evaluate the model's generalization to unseen systematic perturbations, revealed meaningful improvements following adversarial training. While the baseline model trained on the original SNLI dataset achieved 85.0% accuracy, the fine-tuned model attained 87.9%, reflecting a +2.9% gain (~90 examples). These results confirm that adversarial training with LLM-generated contrast sets enhances robustness on both seen perturbations and novel examples, improving systematic generalization.

## 3.4 Error Analysis by Confusion Matrix

The confusion matrices in Figure 1 offer a detailed perspective on how adversarial training improved the model's performance across the original test set, original contrast set, and new contrast set. Key observations are as follows:

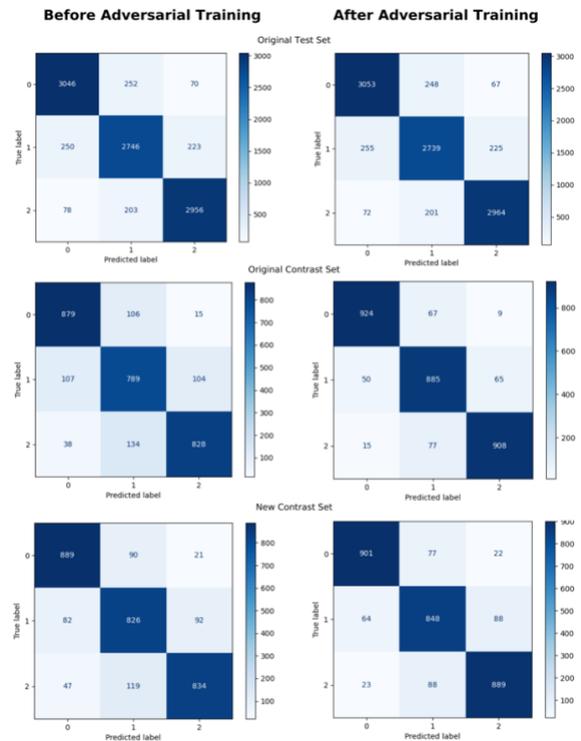

**Figure 1:** Confusion matrices before and after adversarial training

**Original Test Set** Adversarial training preserved the model's performance on standard examples, as indicated by the near-identical diagonal dominance in the confusion matrices. This stability demonstrates that exposure to adversarial data did not degrade the model's ability to classify unperturbed examples. Improvements were observed in previously misclassified neutral examples, although a slight decline occurred in the performance of classifying actual neutral labels. The most notable gains were in resolving misclassifications between entailments and contradictions, with entailments previously classified as contradictions and vice versa showing marked improvement.

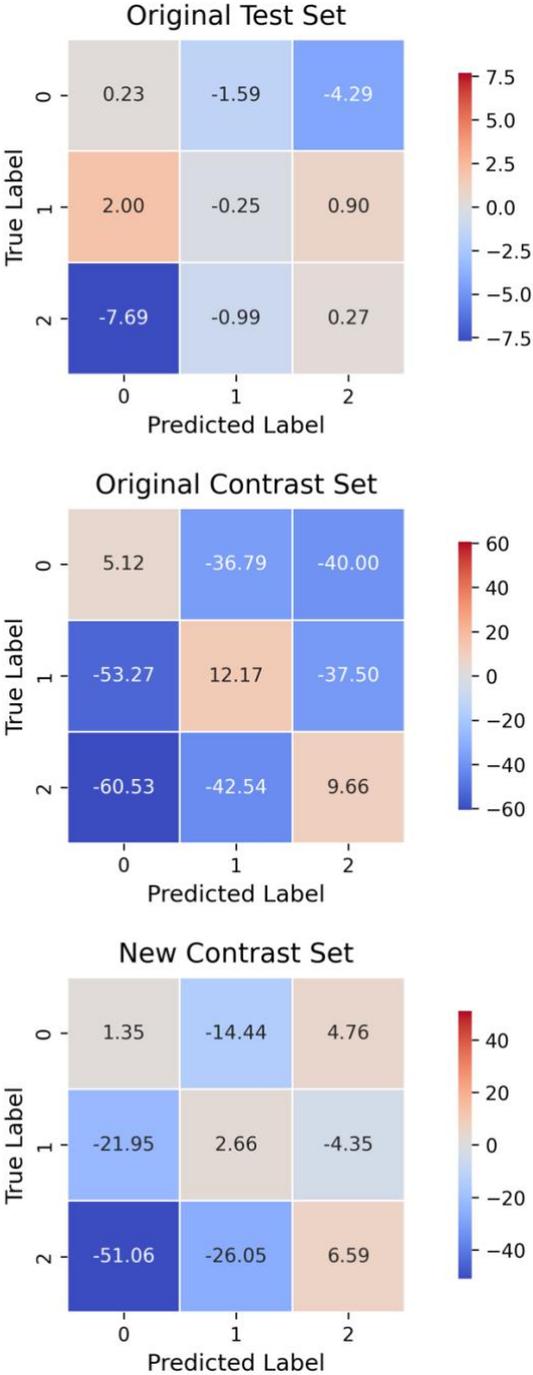

**Figure 2:** Percent change in confusion matrices post-adversarial training

However, these enhancements may be influenced by smaller sample sizes in the confusion matrix's top-right and bottom-left corners, potentially exaggerating the observed gains.

**Original Contrast Set** Adversarial training resulted in substantial increases in correct classifications and reductions in off-diagonal misclassifications. The largest improvements were seen in classifying neutral examples as seen in Figure 3. This refinement suggests that the contrast set effectively pushed the model to better examine more ambiguous cases, improving its understanding of the decision boundaries in this critical area.

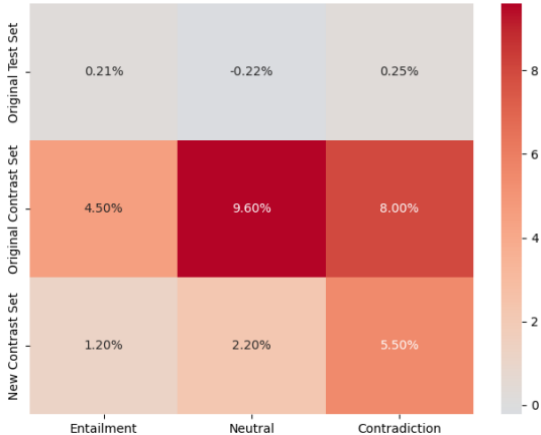

**Figure 3:** Improvements in accuracy by label type

**New Contrast Set** While performance increases on the new contrast set were less pronounced compared to the original contrast set, they were consistently observed across all but one category. The exception was in entailments misclassified as contradictions (Figure 2), where a slight decrease in performance was noted. Interestingly, the largest gains came from contradictions misclassified as entailments, suggesting that training on the original contrast set might have introduced a bias favoring the prediction of contradictions over entailments. However, this trend may also stem from the aforementioned smaller sample sizes in the affected regions of the confusion matrix, making this determination not yet statistically robust.

### 3.5 Error Analysis by Perturbation Type

A detailed analysis of error rates before and after adversarial training reveals significant improvements across all six types of label shifts, as illustrated by the percent change in error rates for both the original and new contrast sets in Figure 4. Key observations are as follows:

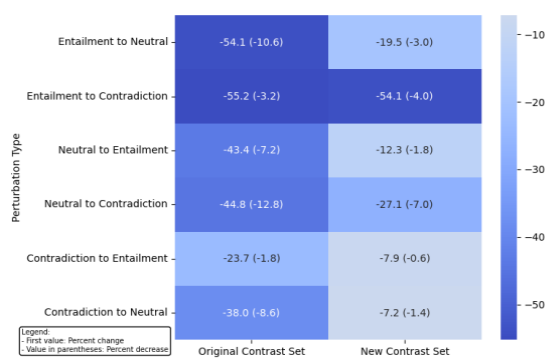

**Figure 4:** Percent change in error rates for each perturbation type

**Original Contrast Set** Adversarial training produced substantial performance gains across all perturbation types, consistent with the increased exposure to systematic perturbations. The largest improvement was observed in entailment-to-contradiction shifts, while the smallest improvement occurred in contradiction-to-entailment shifts. This trend is noteworthy and may suggest that training on the contrast set introduced a slight bias favoring the prediction of contradiction. That said, the statistical significance of this observation is yet to be determined due to smaller sample sizes in these categories.

However, the order of improvement magnitude further supports this hypothesis. Perturbations originating from entailment (e.g., entailment-to-contradiction and entailment-to-neutral) saw the largest reductions in error rates, followed by perturbations involving neutral labels, and finally, perturbations originating from contradiction. This progression likely reflects the model's initial difficulties with entailment-based transitions and its enhanced ability to differentiate these boundaries post-adversarial training.

**New Contrast Set** Improvements on the new contrast set were more moderate compared to the original contrast set but still consistent across all perturbation types. The most pronounced gains were observed for perturbations involving shifts to contradiction, such as neutral-to-contradiction and entailment-to-contradiction. This suggests that the new contrast set effectively refined the model's understanding of decision boundaries adjacent to the contradiction class.

### 3.6 Analysis of Corrected Misclassifications

In the new contrast set, 4.70% of examples were misclassified by the baseline model and correctly classified by the adversarially trained model, with the most significant improvements observed in predictions for the contradiction class (accounting for 44% of the corrections). These gains highlight enhanced logical reasoning and semantic understanding, particularly in recognizing inconsistencies and handling numerical relationships. For instance, the baseline model misclassified the premise, *"Three children hold a boy's arms down while another boy in a hat shoots a water gun at him,"* paired with the hypothesis, *"There are fewer than four children present,"* as entailment, likely due to a superficial interpretation of the premise. In contrast, the adversarially trained model correctly identified it as a contradiction, showcasing improved numerical and logical reasoning. Another example involves the premise, *"Four people and a baby are crossing the street at a crosswalk,"* paired with the hypothesis, *"Fewer than four people and a baby are crossing the street at the crosswalk."* Here, the baseline model predicted neutral, struggling with the quantifiers, while the adversarially trained model accurately classified it as a contradiction. These examples illustrate the adversarially trained model's enhanced ability to interpret numerical relations, handle negations, and recognize subtle logical nuances, leading to better performance after adversarial training on the original contrast set.

### 3.7 Analysis of New Misclassifications

Despite improving overall performance, adversarial training also introduced a small number of new errors, with 1.73% of the new contrast set being misclassified by the adversarially trained model despite being correctly handled by the baseline model. These new errors predominantly fell within the neutral category (accounting for 56% of the errors), which often requires subtle reasoning and balanced judgment. This pattern suggests that adversarial training may have inadvertently fostered overgeneralization or heightened sensitivity to specific patterns, potentially undermining nuanced reasoning. For example, the premise, *"The boys are playing with Legos,"* paired with the hypothesis, *"The boys are *developing their spatial reasoning skills,"* was misclassified as entailment by the adversarially trained model, likely due to overemphasis on implicit associations between activities and their inferred benefits, where neutral would have been more appropriate. Similarly, in the case of the premise, *"Two kids running past a dinosaur in the woods,"* paired with the hypothesis, *"The dinosaur exhibit at the local zoo recently acquired a new animatronic model,"* the adversarially trained model misclassified this as a contradiction. This likely stemmed from over-sensitivity to contextual mismatches, interpreting "woods" and "zoo" as conflicting instead of unrelated. These errors reflect an increased sensitivity to contextual nuances and a potential bias toward definitive labels like entailment or contradiction. Such biases may result from overfitting to adversarial examples and underscore the need for further refinements to the adversarial training process to preserve optimal model reasoning.

## 4 Discussion and Related Works

### 4.1 LLM-Generated Contrast Sets

Contrast sets have become a cornerstone of robustness evaluation in NLP datasets since their introduction by Gardner et al. (2020). These manually curated sets systematically perturb existing datasets to challenge models at decision boundaries, uncovering vulnerabilities that are often masked by strong performance on standard test sets. Gardner et al. (2020) demonstrated that state-of-the-art models perform significantly worse on contrast sets, revealing their reliance on spurious correlations and dataset artifacts. However, while highly effective, human-annotated contrast sets are inherently limited by their manual, labor-intensive nature, which constrains feasibility and scalability.

Automated approaches, such as those proposed by Lie et al. (2020), have sought to address these scalability issues. By employing modular, rule-based transformations of linguistic phenomena, Lie et al. (2020) showcased the feasibility of programmatic, automated contrast set generation. However, this approach faces significant limitations:

**Limited Coverage:** Lie et al. (2020) transformed only 19.7% of SNLI instances, leaving most examples unchanged. This limited coverage reduces the amount of usable data and introduces potential biases, particularly as the ACE/ERG grammar system struggles with more complex sentence structures.

**Dependence on Grammar Systems:** The methodology relies heavily on the ACE/ERG grammar system, a resource that took over 20 years of human labor to develop and is restricted to English. This dependence not only limits the scalability of the approach but also constrains its applicability to other languages. State-of-the-art LLMs mitigate these limitations, enabling broader scalability and linguistic generalization.

**Linguistic Diversity:** While Lie et al. (2020) systematically perturbed data, their transformations lacked the depth necessary for comprehensive model evaluation across a broad range of linguistic phenomena. Their study focused on seven predefined linguistic categories—polar questions, it-clefts, tense and aspect, modality, negation, passives, and subject-object swapping. Any additional phenomena required manual integration into their grammar system before corresponding examples could be generated.

Our study addresses these challenges by leveraging Google Gemini 1.5 Pro to generate contrast sets that achieve full dataset coverage and capture nuanced linguistic transformations. Unlike rule-based systems, our approach avoids reliance on predefined grammatical frameworks, allowing greater flexibility and applicability to a broader range of linguistic phenomena.

Our LLM-generated contrast sets, comprising 3,000 examples each, are 15 times larger than PERSPECTRUM, the dataset most similar to a contrast set for SNLI manually curated by Gardner et al. (2020). It also surpasses the grammar-based approach in Lie et al. (2020) in terms of linguistic diversity. The most notable distinction of our approach however, lies in its unparalleled efficiency: 3,000 examples were generated in just 90 minutes, compared to an average of 3 minutes per example for PERSPECTRUM's human annotators. This translates to generating 99 examples in the same time frame, underscoring our method's practicality for large-scale operations.

### 4.2 Adversarial Training with Contrast Sets

While contrast sets are not inherently adversarial, they similarly compel models to address decision boundary complexities and diverse linguistic phenomena. Studies such as Gardner et al. (2020) and Lie et al. (2020) have shown that contrast sets expose reliance on dataset artifacts and that adversarial training improves performance on out-of-distribution data without compromising standard performance. Building on this, our study fine-tuned the ELECTRA-small model using a dataset combining original SNLI examples with an LLM-generated contrast set. This approach improved accuracy on the contrast set by +7.4%, reducing reliance on spurious correlations, while test set accuracy remained stable with a marginal +0.1% improvement. Generalization to unseen perturbations also increased by +2.9%, highlighting the potential of systematically generated contrast sets to address challenges beyond seen perturbations.

### 4.3 Framework Limitations

While our framework offers clear benefits, several limitations remain. A key challenge is the lack of explicit categorization of linguistic phenomena, such as tense shifts, negations, and modality changes. This limits more detailed error analysis and the development of targeted training strategies. Establishing a systematic taxonomy of linguistic categories could enable more precise evaluations and enhance robustness across specific linguistic challenges.

The effectiveness of our approach also depends heavily on the quality of the LLM used to generate contrast sets. As LLMs continue to improve, so will their ability to produce diverse and accurate perturbations, enabling the method to be more effective and reliable.

## 5 Conclusion

This study addresses the challenges of evaluating robustness and improving systematic generalization in NLP models by using large language models to generate scalable, diverse contrast sets. Using the SNLI dataset, we demonstrated that LLM-generated contrast sets reveal model vulnerabilities and improve robustness when used for adversarial training. Fine-tuning the ELECTRA-small model with these sets significantly enhanced performance on perturbed datasets without compromising standard benchmark accuracy. The method also improved generalization to unseen perturbations. By automating contrast set creation, this approach provides researchers and practitioners with a practical tool to evaluate NLP models and enhance their performance on real-world challenges.

## A Additional Metrics

Here, we present metrics related to, but not directly referenced in the paper.

### A.1 Error Rates Across Perturbations

| Perturbation | Baseline | Post-Training |
| --- | --- | --- |
| Ent. → Neu. | 19.6 % | 9.0 % |
| Ent. → Con. | 5.8 % | 2.6 % |
| Neu. → Ent. | 16.6 % | 9.4 % |
| Neu. → Con. | 28.6 % | 15.8 % |
| Con. → Ent. | 7.6 % | 5.8 % |
| Con. → Neu. | 22.6 % | 14.0 % |

**Table 1:** Error rates on the original contrast set before and after adversarial training

| Perturbation | Baseline | Post-Training |
| --- | --- | --- |
| Ent. → Neu. | 15.4 % | 12.4 % |
| Ent. → Con. | 7.4 % | 3.4 % |
| Neu. → Ent. | 14.6 % | 12.8 % |
| Neu. → Con. | 25.8 % | 18.8 % |
| Con. → Ent. | 7.6 % | 7.0 % |
| Con. → Neu. | 19.4 % | 18.0 % |

**Table 2:** Error rates on the new contrast set before and after adversarial training

## A.2 Accuracy by Label

| Label | Baseline | Post-Training |
|---|---|---|
| Entailment | 90.4 % | 90.6 % |
| Neutral | 85.3 % | 85.1 % |
| Contradiction | 91.3 % | 91.6 % |

Table 3: Accuracy on the original test set before and after adversarial training

| Label | Baseline | Post-Training |
|---|---|---|
| Entailment | 87.9 % | 92.4 % |
| Neutral | 78.9 % | 88.5 % |
| Contradiction | 82.8 % | 90.8 % |

Table 4: Accuracy on the original contrast set before and after adversarial training

| Label | Baseline | Post-Training |
|---|---|---|
| Entailment | 88.9 % | 90.1 % |
| Neutral | 82.6 % | 84.8 % |
| Contradiction | 83.4 % | 88.9 % |

Table 5: Accuracy on the new contrast set before and after adversarial training

## A.3 Baseline vs. Adversarially Trained Model

|  | Adv. Correct | Adv. Incorrect |
|---|---|---|
| **Baseline Correct** | 8558 (87.1 %) | 190 (1.9 %) |
| **Baseline Incorrect** | 198 (2.0 %) | 878 (9.0 %) |

Table 6: Performance comparison on the original test set

|  | Adv. Correct | Adv. Incorrect |
|---|---|---|
| **Baseline Correct** | 2484 (82.8 %) | 12 (0.4 %) |
| **Baseline Incorrect** | 233 (7.8 %) | 271 (9.0 %) |

Table 7: Performance comparison on the original contrast set

|  | Adv. Correct | Adv. Incorrect |
|---|---|---|
| **Baseline Correct** | 2497 (83.2 %) | 52 (1.7 %) |
| **Baseline Incorrect** | 141 (4.7 %) | 310 (10.4 %) |

Table 8: Performance comparison on the new contrast set